\documentclass[twoside,11pt]{article}
\makeatletter
\newlength\titlebox
\twocolumn
\def\addcontentsline#1#2#3{}

\def\maketitle{%
  \par%
  \begingroup%
     \def\thefootnote{\fnsymbol{footnote}}%
     \def\@makefnmark{\rlap{$^{\@thefnmark}$\hss}}%
     \long\def\@makefntext##1{%
                  \parindent 1em\noindent%
                  \hbox to 1em{$^{\@thefnmark}$}##1}
     \twocolumn[\@maketitle] \@thanks%
  \endgroup%
  \setcounter{footnote}{0}%
  \let\maketitle\relax\let\@maketitle\relax%
  \gdef\@thanks{}\gdef\@author{}\gdef\@title{}%
  \let\thanks\relax}

\def\@maketitle{%
  \vbox to \titlebox{%
    \hsize\textwidth\linewidth\hsize%
    \vskip 0.125in minus 0.05in%
    \centering{\Large\bf \@title \par}%
    \vskip 0.2in plus 0.1fil minus 0.1in
    {\def\and{\unskip\enspace{\rm and}\enspace}%
     \def\And{\end{tabular}\hss \egroup \hskip 1in plus 2fil 
              \hbox to 0pt\bgroup\hss \begin{tabular}[t]{c}\bf}%
     \def\AND{\end{tabular}\hss\egroup \hfil\hfil\egroup
              \vskip 0.25in plus 1fil minus 0.125in
              \hbox to \linewidth\bgroup\large \hfil\hfil
              \hbox to 0pt\bgroup\hss \begin{tabular}[t]{c}\bf}
    \hbox to \linewidth \bgroup\large \hfil\hfil
    \hbox to 0pt\bgroup\hss \begin{tabular}[t]{c}\bf\@author 
                            \end{tabular}\hss\egroup
    \hfil\hfil\egroup}
  \vskip 0.3in plus 2fil minus 0.1in
}}

\renewenvironment{abstract}{\section*{\centerline{Abstract}}}{}

\def\section{%
    \@startsection{section}{1}{\z@}%
                  {-2.0ex plus -0.5ex minus -0.3ex}%
                  {0.8ex plus 0.3ex minus 0.2ex}%
                  {\large\bf\raggedright}}
\def\subsection{%
    \@startsection{subsection}{2}{\z@}%
                  {-1.4ex plus -0.4ex minus -0.2ex}%
                  {0.6ex plus 0.2ex minus 0.1ex}%
                  {\normalsize\bf\raggedright}}
\def\subsubsection{%
    \@startsection{subsubsection}{3}{\z@}%
                  {-0.8ex plus -0.3ex minus -0.1ex}%
                  {0.4ex plus 0.1ex minus 0.1ex}%
                  {\normalsize\bf\raggedright}}
\def\paragraph{%
    \@startsection{paragraph}{4}{\z@}%
                  {-0.8ex plus -0.3ex minus -0.1ex}%
                  {-1em}%
                  {\normalsize\bf}}
\def\subparagraph{%
    \@startsection{subparagraph}{5}{\parindent}%
                  {0.4ex plus 0.3ex minus 0.1ex}%
                  {-1em}%
                  {\normalsize\bf}}

\labelwidth\leftmargini\advance\labelwidth-\labelsep \labelsep 5pt

\ifcase\@ptsize%
    \renewcommand{\normalsize}{
        \@setsize\normalsize{11.3pt}\xpt\@xpt%
        \abovedisplayskip 10\p@\@plus2\p@\@minus5\p@%
        \abovedisplayshortskip\z@\@plus3\p@%
        \belowdisplayshortskip 4\p@\@plus3\p@\@minus3\p@%
        \belowdisplayskip\abovedisplayskip%
        \let\@listi\@listI}%
 \or%
    \renewcommand{\normalsize}{
        \@setsize\normalsize{12.6pt}\xipt\@xipt%
        \abovedisplayskip11\p@\@plus2\p@\@minus4\p@%
        \abovedisplayshortskip\z@\@plus3\p@%
        \belowdisplayshortskip5\p@\@plus3\p@\@minus2\p@%
        \belowdisplayskip\abovedisplayskip%
        \let\@listi\@listI}%
 \or%
    \renewcommand{\normalsize}{
        \@setsize\normalsize{13pt}\xiipt\@xiipt%
        \abovedisplayskip 11\p@ \@plus3\p@ \@minus5\p@%
        \abovedisplayshortskip \z@ \@plus3\p@%
        \belowdisplayshortskip 5\p@ \@plus3\p@ \@minus2\p@%
        \belowdisplayskip\abovedisplayskip%
        \let\@listi\@listI}%
 \fi    

\ifcase\@ptsize%
    \renewcommand{\small}{
        \@setsize\small{10.5pt}\ixpt\@ixpt%
        \abovedisplayskip 8\p@ \@plus3\p@ \@minus3\p@%
        \abovedisplayshortskip \z@ \@plus2\p@%
        \belowdisplayshortskip 3\p@ \@plus2\p@ \@minus2\p@%
        \belowdisplayskip\abovedisplayskip%
        \def\@listi{\leftmargin\leftmargini%
                    \topsep 3.5\p@ \@plus1.5\p@ \@minus1.5\p@%
                    \parsep 1.5\p@ \@plus\p@ \@minus\p@%
                    \itemsep \parsep}}%
 \or%
    \renewcommand{\small}{
        \@setsize\small{11.3pt}\xpt\@xpt%
        \abovedisplayskip 9\p@ \@plus2\p@ \@minus4\p@%
        \abovedisplayshortskip \z@ \@plus3\p@%
        \belowdisplayshortskip 5\p@ \@plus2.5\p@ \@minus2.5\p@%
        \belowdisplayskip\abovedisplayskip%
        \def\@listi{\leftmargin\leftmargini%
                    \topsep 5\p@ \@plus2\p@ \@minus2\p@%
                    \parsep 2\p@ \@plus2\p@ \@minus\p@%
                    \itemsep \parsep}}%
 \or%
    \renewcommand{\small}{
        \@setsize\small{12pt}\xipt\@xipt%
        \abovedisplayskip 9\p@ \@plus3\p@ \@minus4\p@%
        \abovedisplayshortskip \z@ \@plus3\p@%
        \belowdisplayshortskip 5\p@ \@plus2.5\p@ \@minus2\p@%
        \belowdisplayskip\abovedisplayskip%
        \def\@listi{\leftmargin\leftmargini%
                    \topsep 5.5\p@ \@plus2.5\p@ \@minus2.5\p@%
                    \parsep 4\p@ \@plus1.5\p@ \@minus\p@%
                    \itemsep \parsep}}%
 \fi

\ifcase\@ptsize
    \renewcommand{\footnotesize}{
        \@setsize\footnotesize{9.3pt}\viiipt\@viiipt%
        \abovedisplayskip 5\p@ \@plus2\p@ \@minus3\p@%
        \abovedisplayshortskip \z@ \@plus\p@%
        \belowdisplayshortskip 2.5\p@\@plus\p@\@minus2\p@%
        \belowdisplayskip\abovedisplayskip%
        \def\@listi{\leftmargin\leftmargini%
                    \topsep 2.5\p@ \@plus\p@ \@minus\p@%
                    \parsep 1.5\p@ \@plus\p@ \@minus\p@%
                    \itemsep \parsep}}%
 \or%
    \renewcommand{\footnotesize}{
        \@setsize\footnotesize{10.3pt}\ixpt\@ixpt%
        \abovedisplayskip 7\p@ \@plus2\p@ \@minus4\p@%
        \abovedisplayshortskip \z@ \@plus\p@%
        \belowdisplayshortskip 3\p@ \@plus2\p@ \@minus2\p@%
        \belowdisplayskip\abovedisplayskip%
        \def\@listi{\leftmargin\leftmargini%
                    \topsep 3\p@ \@plus2\p@ \@minus2\p@%
                    \parsep 2\p@ \@plus\p@ \@minus\p@%
                    \itemsep \parsep}}%
 \or%
    \renewcommand{\footnotesize}{
        \@setsize\footnotesize{11pt}\xpt\@xpt%
        \abovedisplayskip 9\p@ \@plus2\p@ \@minus4\p@%
        \abovedisplayshortskip \z@ \@plus3\p@%
        \belowdisplayshortskip 5\p@ \@plus3\p@ \@minus3\p@%
        \belowdisplayskip\abovedisplayskip%
        \def\@listi{\leftmargin\leftmargini%
                    \topsep 4.5\p@ \@plus2\p@ \@minus2\p@%
                    \parsep 3\p@ \@plus\p@ \@minus\p@%
                    \itemsep \parsep}}%
 \fi

\ifcase\@ptsize%
    \renewcommand{\large}{\@setsize\large{13pt}\xiipt\@xiipt}
 \or%
    \renewcommand{\large}{\@setsize\large{13pt}\xiipt\@xiipt}
 \or%
    \renewcommand{\large}{\@setsize\large{16pt}\xivpt\@xivpt}
 \fi

\ifcase\@ptsize%
    \renewcommand{\Large}{\@setsize\Large{16pt}\xivpt\@xivpt}
 \or%
    \renewcommand{\Large}{\@setsize\Large{16pt}\xivpt\@xivpt}
 \or%
    \renewcommand{\Large}{\@setsize\Large{16pt}\xivpt\@xivpt}
 \fi

\ifcase\@ptsize%
 \or%
 \or%
 \fi

\ifcase\@ptsize%
    \def\@listi{\leftmargin\leftmargini
                \topsep  6\p@ \@plus2\p@ \@minus2\p@%
                \parsep  2\p@ \@plus0.5\p@ \@minus\p@%
                \itemsep 2.5\p@ \@plus\p@ \@minus0.5\p@}%
 \or%
    \def\@listi{\leftmargin\leftmargini
                \topsep  8\p@ \@plus2\p@ \@minus2\p@%
                \parsep  3\p@ \@plus1.5\p@ \@minus\p@%
                \itemsep 3\p@ \@plus1.5\p@ \@minus\p@}%
 \or%
    \def\@listi{\leftmargin\leftmargini
                \topsep  9\p@ \@plus3\p@   \@minus4\p@%
                \parsep  4\p@  \@plus2\p@ \@minus\p@%
                \itemsep 4\p@  \@plus2\p@ \@minus\p@}%
 \fi
\let\@listI\@listi

\makeatother

\usepackage{avm,times,chicago,amsmath,amssymb,covingtn,ipa} 
\usepackage{vmargin}
\setpapersize{A4}
\setmargins{2.2cm}{3.4cm}{165mm}{230mm}{0pt}{0pt}{0pt}{11mm}

\newcommand{\finalpagebreak}{\pagebreak}
\newcommand{\finalforcedpage}{\enlargethispage*{100cm}}

\normalsize

\avmhskip{.2em}
\avmvskip{.05ex}
\avmfont{\footnotesize\sf}
\avmsortfont{\footnotesize\it}
\avmvalfont{\footnotesize\it}
\avmoptions{center}

\newcommand{\Star}{$\ast$}
\newcommand{\lingnull}{$\varnothing${}}
\renewcommand{\root}[1]{\raisebox{1mm}{\mbox{$\sqrt{\,}$}}$#1$\/}
\newcommand{\url}[1]{\footnotesize\texttt{#1}}
\newcommand{\code}[1]{{\small\tt #1}}
\newcommand{\strich}{\rule{.99\linewidth}{.2pt}\\}
\newcommand{\startpiece}{\noindent\strich\vspace{-\baselineskip}\footnotesize}
\newcommand{\stoppiece}{\vspace{-\baselineskip}\noindent\strich\normalsize}
\title{
Computing Declarative Prosodic Morphology}
\author{Markus Walther \\
  Seminar f\"ur Allgemeine Sprachwissenschaft \\ Heinrich-Heine-Universit\"at D\"usseldorf \\
Universit\"atsstr. 1, D-40225 D\"usseldorf, Germany \\
\texttt{walther@ling.uni-duesseldorf.de}}
\date{}
\begin{document}
\maketitle
\begin{abstract}
This paper describes a computational, declarative approach
to prosodic morphology that uses inviolable constraints to denote small finite candidate sets
which are filtered by a restrictive incremental optimization mechanism.
The new approach is illustrated with an implemented fragment of Modern
Hebrew verbs couched in MicroCUF, an expressive
constraint logic formalism. For  generation {\em and} parsing of word forms,
I propose a novel off-line technique to eliminate run-time optimization. It produces a
finite-state oracle that efficiently restricts the constraint
interpreter's search space. As a byproduct, unknown words can be analyzed without special
mechanisms. Unlike pure finite-state transducer approaches, 
this hybrid setup allows for more expressivity in constraints to specify
e.g.  token identity for reduplication or arithmetic constraints for
phonetics.
\end{abstract}
\section{Introduction}
\label{sec:intro}
Prosodic morphology (PM) circumscribes a number of phenomena ranging from
`nonconatenative' root-and-pattern morphology over infixation 
to various cases of
reduplication, where the phonology strongly influences the shape of words by way
of obedience to structural constraints defining wellformed morae, syllables, feet
etc. These phenomena have been difficult to handle in
earlier rule-based treatments \cite[159 ff.]{sproat:92}. 
Moreover, as early as \citeN{kisseberth:70} authors have noted that
derivational accounts of PM are bound to miss important linguistic generalizations that are best
expressed via constraints. Kisseberth showed that verb stems in
Tonkawa, a Coahuiltecan language,                        
display a complex V/\lingnull{} alternation pattern when various affixes
are added (fig. \ref{fig:tonkawa}).
\begin{figure}[htb]
\begin{tabular}[t]{ll@{}r@{}}
%
`{\em to cut}'  & `{\em to lick}'  
\\ \hline
picn-o\glotstop      & netl-o\glotstop  & 
({\footnotesize\it 3sg.obj.stem-3sg.subj.}) \\  
%
we-pcen-o\glotstop   & we-ntal-o\glotstop & 
({\footnotesize\it 3pl.obj.-stem-3sg.subj.}) \\
%
picna-n-o\glotstop  & netle-n-o\glotstop  &  
({\footnotesize\it 3sg.obj.stem-prog.
-3sg.subj.})  \\ \hline
%
p(i)c(e)n(a)   & n(e)t(a)l(e)  & {\hfill\footnotesize\it stems \hfill}
\end{tabular}
\caption{Tonkawa verb forms with V/\lingnull{} effects}
\label{fig:tonkawa}
\vspace{-3\baselineskip}
\end{figure}
 This leads to more and more complicated vowel deletion rules as
the fragment is enlarged. In contrast, a straightforward constraint that
bans three consecutive consonants offers a unified account of the conditions
under which vowels must surface. Later developments have refined constraints
such as \Star CCC to refer to syllable structure instead: complex
codas and onsets are disallowed. At least since
\citeN{kahn:76}, \citeN{selkirk:82}, such segment-independent reference
to syllable structure has been standardly assumed in the generative
literature.

Astonishing as it may be, even the latest computational models of PM
phenomena apparently eschew the incorporation of real prosodic
representations, syllabification and constraints. \citeN{kiraz:96c} uses multi-tape
two-level morphology to analyze some Arabic data, but -- despite the
suggestive title -- must simulate
prosodic operations such as `add a mora' by their extensionalized
rule counterparts, which refer to C or V segments instead of
moras. There is no on-line syllabification and the
exclusive use of lexically prespecified syllable-like symbols on a
separate templatic pattern tape renders his approach vulnerable to 
postlexical resyllabification effects. 
Similarly, \citeN{beesley:96} seems content in employing a great number of
CV templates in his large-scale finite-state 
model of Arabic morphology, which are intersected with lexical roots
and then transformed to surface realizations by various epenthesis,
deletion and assimilation rules. Beesley states that further application of
his approach to e.g. Hebrew is foreseen. On the downside, however,
again there is no real prosody in his model; the relationship between  
template form and prosody is not captured.

Optimality Theory (OT, \citeNP{prince.smolensky:93}), as applied to PM
\cite{mccarthy.prince:93}, {\em does} claim to capture this
relationship, using a ranked set of violable prosodic constraints
together with global violation minimization.  However, to date there 
exist no sufficiently formalized analyses of nontrivial PM
fragments that could be turned into testable computational models.
The OT framework itself has been shown to be expressible with weighted
finite-state automata, weighted intersection and best-path
algorithms \cite{ellison:94} if constraints and OT's
GEN component --
the function from underlying forms to prosodified surface forms -- are
regular sets. A recent proposal by \citeN{karttunen:98} dispenses
with the weights while still relying on the same regularity assumption. Published PM
analyses, however, frequently make use of 
constraint parametrizations from the ALIGN family, which requires greater
than regular power \cite{ellison:95a}. Further developments of 
 OT such as correspondence theory 
--
 extensively used in much newer work on PM -- have not received a formal
 analysis so far. Finally, although OT postulates that constraints are
 universal, this metaconstraint has been violated from the outset,
 e.g. in presenting Tagalog {\em -um-}  as a language-specific
 parameter to ALIGN in \citeN{prince.smolensky:93}. Due to
 the convincing presentation of a number of other
 forceful arguments against constraint universality in  \citeN{ellison:97}, the case for
 language-specific constraints must clearly be seen as reopened, and -- as a
 corollary -- the case for constraint inviolability as well.

Declarative Phonology (DP, \citeNP{bird:95}, \citeNP{scobbie:91}) is
just such a constraint-based framework that dispenses 
with violability and requires a monostratal conception of phonological
grammar, as compared to the multi-level approaches 
discussed above. 
Both abstract generalizations and concrete morphemes are 
expressed by constraints.
DP requires analyses to be formally adequate, i.e. use a grammar description
language with formal syntax and semantics. As a consequence,
Chomsky's criteria for a generative grammar which must be
 ``perfectly explicit'' and
``not rely on the intelligence of the understanding reader''
\cite[4]{chomsky:65} are automatically fulfilled.
DP thus appears to be a good starting point for a restrictive,
surface-true theory of PM that is explicitly computational.

The rest of this paper reviews in informal terms  the theory of
\citeN{walther:97} (section \ref{sec:DPM}), showing in formal detail
in section \ref{sec:imp_ana} how to implement a concrete analysis of Modern Hebrew verbs. 
Section \ref{sec:pargen} explains a novel approach to both generation {\em
  and} parsing of word forms under the new theory.
The paper concludes in section \ref{sec:disc}.

\section{Declarative Prosodic Morphology}
\label{sec:DPM}
Focussing on cases of `nonconcatenative' root-and-pattern morphology, Declarative
Prosodic Morphology (DPM) starts with an intuition that is 
opposite to what the traditional idea of templates or fixed phonological shapes
\cite{mccarthy:79} suggests,
namely that {\em shape variance} is actually quite common and should
form the analytical 
basis for theoretical accounts of PM. Besides the Tonkawa case
(fig.\ref{fig:tonkawa}), shape variance is also at work in Modern
Hebrew (MH) inflected verb forms \cite{glinert:89}, see
fig. \ref{fig:hebrew}.%
\footnote{Regular MH verbs are traditionally divided into seven verbal classes
  or {\em binyanim}, B1-B7. Except for B4 and B6, which regularly act as
  passive counterparts of B3 and B4, the semantic contribution of each
  class is no longer transparent in the modern language. Also,
  in many cases the {\em root} (written \root{C_1.C_2.C_3}) is
  restricted to an idiosyncratic subset of the binyanim.

  An a-templatic treatment of MH prosodic morphology was
  first proposed by \citeN[40ff.]{bat-el:89} within an unformalized,
  non-surface-true,  non-constraint-based setting.}
\begin{figure}[htb]
\begin{center}
\begin{tabular}[t]{@{}ll|ll@{}}
 & past & future \\ \hline
\footnotesize\it 1sg &          gamar-ti        & \glotstop e-gmor \\
\footnotesize\it 2sg.m &       gamar-ta        & ti-gmor       \\
\footnotesize\it 2sg.f  &       gamar-t         & ti-gmer-i     \\
\footnotesize\it 3sg.m &               gamar           & ji-gmor       \\
\footnotesize\it 3sg.f       &               gamr-a          & ti-gmor       \\
\footnotesize\it 1pl       &               gamar-nu        & ni-gmor       \\
\footnotesize\it 2pl      &               gamar-tem       & ti-gmer-u     \\
\footnotesize\it 3pl      &               gamr-u          & ji-gmer-u
\end{tabular}
\end{center}
\caption{Modern Hebrew \root{g.m.r} {\em `finish'} (B1)}
\label{fig:hebrew}
\end{figure}
Here we see a systematic V/\lingnull{} alternation
of both stem vowels, depending on the affixation pattern. This results
in three stem shapes {\em CVCVC}, {\em CVCC} and {\em CCVC}. Any
analysis that simply stipulates shape selection on the basis of specific
inflectional categories {\em or} phonological context (e.g. 3sg.f
$\vee$ 3pl {\em or}
-V $\rightarrow$ CVCC$_{stem}$ / \underline{B1 past}) misses the fact
that the shapes, their alternating behaviour and their proper
selection are derivable.
Derivational repairs by means of `doubly open syllable' syncope rules
(/ga.\underline{ma.r-a.}/ $\rightarrow$ /.gam.ra./)
are similarly {\em ad hoc}.

$\bullet$ A first  step in developing an alternative DPM analysis of MH verbs is to
explicitly recognize {\bf alternation} of an element X {\bf with zero} --
informally written (X) -- as a
serious formal device besides its function as a piece of merely descriptive notation (cf.
\citeNP{hudson:86} for an earlier application to Arabic).
In contrast to nonmonotonic deletion or epenthesis, (X) is
a surface-true declarative expression \cite[93f.]{bird:95}. The reader is reminded that
DP sees grammar expressions as partial formal {\em descriptions of
  sets} of phonological 
objects. The former reside on a different ontological level from the
latter, in contrast to traditional object-to-object transformations on
the same level. Hence a preliminary  grammar expression
$g(V_1)m(V_2)r$ for a
Hebrew stem  (with abstract stem vowels) denotes
 the {\em set} $\{gmr, gV_1mr,
gmV_2r,gV_1mV_2r\}$. Note that the (X) property as attributed to
segmental positions is distinctive -- in contrast to stem vowels root segments do
not normally alternate with zero, and neither do affix segments in an important asymmetry
with stems. This point is reinforced by the exceptions that do exist:
phonologically unpredictable C/\lingnull{} alternation occurs in some
MH stems, e.g.  {\em  natan/laka\nichi} `he gave/took' vs {\em ji-ten /ji-ka\nichi} `he
will give/take'; by surface-true {\em
  (n/l)} encoding we can avoid diacritical solutions here. 

$\bullet$ Step two uses {\bf concatenation} to combine individual {\em descriptions} of stems and
affixes, besides 
connecting segmental positions {\em within}
these linguistic entities. Since, as we have just seen, a single description can denote
several objects of varying surface string length, concatenation (\^{}) at the
description level is actually powerful enough to describe
`nonconcatenative' morphological phenomena. In DPM these do not receive
independent ontological status (cf. \citeNP{bird.klein:90} and
\citeNP{gafos:95} for other formal  and articulatory-phonological arguments
leading to the same conclusion). A more detailed description of the 
3pl.fut. inflected form of \root{g.m.r} might therefore be $j\text{\^{}} i\text{\^{}} g\text{\^{}}
(V_1)\text{\^{}} m\text{\^{}} (V_2)r\text{\^{}} u$. In order to allow
for paradigmatic%
\footnote{See \citeN{walther:97} for a discussion of various ways to
  derive rather than stipulate the {\em syntagmatic} pattern of alternating
  and non-alternating segmental positions within stems.}
generalizations over independent entities such as root 
and stem vowel pattern {\em within} concatenated descriptions, a
hierarchical lexicon conception based on multiple inheritance of named
abstractions can be used (cf. \citeNP{riehemann:93}).

$\bullet$ Step three conjoins a word form description with {\bf declarative
syllabification  and syllable
structure constraints} in order to  
impose prosodic wellformedness conditions. 
For Modern Hebrew (and Tonkawa), the syllable canon 
is basically CV(C). Expressed in prosodic terms,  complex
codas and onsets are banned, while an onset must precede each syllable
nucleus. These
syllable roles are established in the first place by syllabification
constraints that exploit local sonority differences 
between successive segments \cite{walther:93}. Alltogether,
the ensemble of prosodic constraints indeed succeeds in narrowing 
down the set for the 3sg.m past tense form to $\{*.gmr.,
*.gamr., *.gmar., !.ga.mar.\} = /gamar/$. For 3pl. future tense B1,
however, an unresolved ambiguity remains: in $\{.jig.me.ru., .ji.gam.ru.\}$, 
only the first element is grammatical.%
\footnote{Note that the prosodic view explains the pronounced
  influence of (C)V affixes on the shape of the whole word: they
  provide a nonalternating syllable nucleus  which can host 
  adjacent stem consonants. 
}
An important observation is that in general there can be no purely phonological constraint to
disambiguate this type of situation. The reason lies in the 
existence of minimal pairs with different category. In our case,
homophonous /.ji.gam.ru./
is grammatical as 3pl. fut. {\em B2} `they will be finished'. We will
return to the analysis of such cases 
after 
proposing a specific disambiguation mechanism in the next step.

$\bullet$ Step four eliminates the remaining ambiguity by invoking an
{\bf Incremental
Optimization Principle} (IOP): ``For all (X) elements, prefer
the zero alternant as early as possible''. ``Early'' corresponds to
traditional left-to-right directionality, but is meant to be understood w.r.t.
the speech production time arrow. ``As possible'' means that IOP
application to a (X) position nevertheless realizes
X if its omission would lead to a constraint conflict. 
Hence, the IOP correctly rules out the second element of
$\{.jig.me.ru., *.ji.gam.ru.\}$. This is because
$.ji.gam.ru.$ represents a missed chance to leave out /a/, the earlier one of the two stem vowels.
The reader may verify that the IOP as it stands also accounts for the
Tonkawa data of fig. \ref{fig:tonkawa}. Tonkawa lends even clearer
support to IOP's left-to-right nature due to the larger number of V/\lingnull{}
vowels involved. As a 
limiting case, the IOP predicts
the possibility of 
vowelless surface stems,
  e.g. formed by two root consonants combined with vowel-final
  prefix and suffix. This prediction is strikingly confirmed by MH
  forms like {\em te-l\nichi-i} `you (sg.f.) will go' \mbox{\root{(h).l.\text{\nichi}},}
  {\em ti-kn-u} `you/they (pl.) will buy' \root{k.n.\varnothing}, {\em
    ti-tn-i} `you
  (sg.f.) will give' \mbox{\root{(n).t.n};} similar cases exist in Tigrinya. There can be no
  meaningful prosodic characterization of isolated $CC$ {\em stem} shapes; 
  only a wordform-based theory like the present one may explain why these forms exist.

Note that, conceptually, IOP is piggybacked on autonomous
DP-style constraint interaction. It merely filters the small finite set
of objects described by the conjunction of all constraints. From
another angle, IOP can be seen as a single context-free substitute for
the various syncope rules employed in former transformational analyses. The claim is
that fixed-directionality-IOP is the only such mechanism needed to
account for PM phenomena. 

A distinguishing feature of the IOP is its potential for an economical
procedural implementation in incremental production. If 
constraint contexts are sufficiently local, the principle can locally decide over (X)
nonrealizations and there will be very limited backtracking through delayed
detection of constraint violation. Because the IOP stops after finding 
the {\em first} (X) realization pattern that violates no constraints, it has
less formal power than global optimization
which must always consider {\em all} candidates. 
Moreover, the IOP supports economic communication, as it leads to shortest
surface forms wherever possible. Finally, at least for
root-and-pattern morphologies it can be argued to aid in speech perception
as well. This is because the closed 
class of stem vowel patterns is less informative than open-class root
segments. Since IOP-guided vowel omission causes root segments to
(statistically) appear at an earlier point in time 
from the acoustic onset of the word, the IOP hypothesis
actively prunes the size of the cohort of 
competing lexical candidates. As a result, unambigous
recognition will generally be achieved more quickly during continous lexical access. 
In sum, the IOP hypothesis not only possesses overall psycholinguistic
plausibility but actually gives some processing advantage to shape
variance. If future research provides the necessary experimental
confirmation, we have yet another case of performance shaping
competence.  

$\bullet$ Step five returns to the minimal pairs problem highlighted in step
three: what to do with anti-IOP realizations such as  that of /a/ in /.ji.gam.ru./ for B2 fut.?
The answer is {\bf (prosodic) prespecification}. A surface-true
constraint demands that B2 future and infinitive as
well as all of B3, B4 must have an onset role for the first stem
element. Thus, the possibility of IOP eliminating the first stem vowel is
blocked by the constraint inconsistency that arises for the first stem element:
either syllabification licenses an incompatible coda or
first and second stem segment together form an illformed onset
cluster. Note that if the constraint is lexicalized as part
of the grammatical description of first stem position, it will
have a maximally local context, referring to just the position
itself. In general, DPM analyses pay much 
attention to proper attachment sites of constraints in order to
maximize their locality. 

The MH verbal suffix {\em -et} (fem.sg.pres.) illustrates that sometimes
another, segmental mode of prespecification is useful. This suffix
is always preceded by a syllable ending in /e/, although IOP
application alone would e.g. prefer */gom.ret/ over /go.me.ret/ `she
finishes'. The effect is morpheme-specific since other \mbox{\em -VC} suffixes
behave as expected here: {\em gomr-im/ot} `they (masc./fem.) finish'.
One solution is to let part of the suffix definition be a constraint
statement which demands that the segment two positions to
its left must be a front vowel. This move captures both the stability and the
quality of this vowel at the same time. (Apophony constraints ensure that the
second stem vowel is never /i/ except in B5, which significantly has a
different suffix {\em -a} in place of {\em -et}). Note that prespecifying
the presuffixal segment to be in an onset position would not work.
\section{On implementing analyses}
\label{sec:imp_ana}
In the following I show how to implement a toy fragment of MH verbs
using the MicroCUF formalism, a typed, feature-based constraint-logic programming
language suitable for natural language modelling. MicroCUF implements
a subset of CUF \cite{doerre.dorna:93},  inheriting its 
formal semantics. It was initially delevoped by the author to overcome
efficiency problems with CUF's original type system. Additionally, its simpler
implemenation provides an open platform for 
experimental modifications, as needed e.g. for parsing and
generation with DPM. After briefly introducing the essentials of 
MicroCUF first, the MH analysis is developed and explained.

\subsection{The MicroCUF constraint formalism}
\label{sec:microcuf}
This section assumes a basic knowledge of Prolog. Like in Prolog,
MicroCUF variables start with uppercase letters or {\tt \_ }, whereas
relational symbols, features and simplex types start
in lowercase; {\tt \%} marks a comment
(fig. \ref{fig:microcuf}a).
 Relations like \code{member}
are written in functional notation,
with a notationally distinguished result argument on the righthand side of {\tt :=}
and the relation symbol plus its (optional) arguments on the lefthand
side. Subgoals like \code{member(Elem)} can
occur anywhere as subterms.
Instead of Prolog's fixed-arity first order terms, MicroCUF has typed feature terms as its
basic data structures. As illustrated in fig. \ref{fig:microcuf}b, subterms are explicitly
conjoined with {\bf\tt \&} or disjunctively combined with {\bf\tt ;}, while
only type terms may be prefixed by the negation operator $\bf\sim$.
Features like \code{left, cat} are separated from their righthand
value terms by {\bf\tt :}. 
Terms may be
tagged by conjunction with a variable (\code{V1}), allowing for the expression of structure
sharing through multiple occurences of the same
variable. 
Feature appropriateness declarations ({\bf\tt ::}) ensure that both the
term in which a feature occurs and its value are typed.
For comparison, the result value of \code{fs} appears in HPSG-style notation
under fig. \ref{fig:microcuf}c.

\begin{figure}[htb]
\begin{minipage}{\linewidth}
\vspace{\baselineskip}
a.
\footnotesize
\begin{verbatim}
% MicroCUF
member(Elem) := [Elem|_].
member(Elem) := [_|member(Elem)].

%Prolog
member(Elem,[Elem|_]).
member(Elem,[_|Rest]) :- member(Elem,Rest).

\end{verbatim}
\end{minipage}
%

\begin{minipage}{.5\linewidth}
b.
{\footnotesize
\begin{verbatim}
fs:=cat:(~((b2;b3;b5)&past)&V1)&left:cat:V1.
phonlist::[cat:categories].
\end{verbatim}}
c.

\mbox{\code{fs $\equiv\;$}}%
\begin{avm}
\[\it phonlist \\
cat & \@1 \it $\;\neg$ ( (b2 $\vee$ b3 $\vee$ b5) $\wedge$ past ) \\
left & \[ \it phonlist \\
           cat & \@1
         \]
\]
\end{avm}

\end{minipage}
\caption{MicroCUF vs Prolog and HPSG notation}
\label{fig:microcuf}
\end{figure}
%
\subsection{Modern Hebrew verbs in MicroCUF}
\label{sec:mhanalysis}
Below I present a concrete MicroCUF grammar in successive pieces. It encodes a toy fragment
of MH verbs and represents a simplified excerpt from a much larger
computational grammar. 
For lack of space, the type hierarchy -- specifying syllable roles, segments,
morphological categories and word-peripheral position -- and the
definition of \code{syllabify} (formalized in \citeNP{walther:95}) have been omitted.

Let us start the explanation with a basic \code{conc}atenation relation which adds a
position \code{Self} in front of some string of \code{Segments}
(1-6). 

\startpiece
\begin{verbatim}
1 conc(Self, Segments) := 
2    Self & 
3    right:(Segments&left:Self&cat:Cat) &
4    cat:Cat &        
5    classify_position_in_word & 
6    constraints.
7
8 classify_position_in_word := 
9  right:self:'-ini' & left:self:'-fin'.
\end{verbatim}
\stoppiece
Here, the familiar recursive first-rest encoding of lists translates
into \code{self}-\code{right} features. This alone makes self and (arbitrarily long) right-context
references possible. To support looking one or more segmental
positions to the left -- a frequent situation in phonological contexts --  we
supplement it with a new feature \code{left} to yield bidirectional lists. For this doubly-linked
list encoding to be wellbehaved, a step right followed by a step left 
is constrained to return to the same position \code{Self} (3),
thus yielding {\em cyclic} feature structures. Next, the value of the feature \code{cat}
at the current position is connected with its right neighbour (3-4). In
the face of our recursively structured lists this makes morphological
and other global categorial information locally accessible at each
segmental position. Finally, relations to incrementally classify each
segmental position as word-initial, medial or wordfinal and to impose
prosodic constraints are added in (5-6). 

Basic concatenation is used in (10-12) to define X/\lingnull{}
positions. 

\startpiece
\begin{verbatim}
10 x_0(_, Segments) := Segments.
11 x_0(X, Segments) := mark:marked & 
12  conc(X, Segments).
13 
14 obl(X, Segments) := mark:unmarked &
15  conc(X, Segments).
16 
17 is(Segment) := self:seg:Segment.
\end{verbatim}
\stoppiece
The first clause of \code{x\_0} (10) realizes the zero
alternant by equating in its second argument the \code{Segments} to follow 
with the result argument; the first argument holding \code{X} is
unused. It gets used in the second clause (11-12), however, where it is
prefixed to the following \code{Segments} by ordinary concatenation. The value of an
additional feature \code{mark}
specifies that realizing an X position is \code{marked} w.r.t. the
IOP, whereas no such value is prescribed in the first clause. Instead, 
the marking there will be supplied later by adjacent instances of either the second
\code{x\_0} clause or  \code{ obl}
(14-15). The latter is the version of concatenation used for 
specifying {\em obligatory}, i.e. nonalternating positions, which consequently are specified as
\code{unmarked}. Alltogether these means yield fully specified strings w.r.t. markedness
information. We will see below how this simplifies an implementation of the IOP.

As can be seen in the accessor relation \code{is} (17),
phonological segments are actually embedded under a further
feature \code{seg}. This treatment enables structure-sharing of segments
independent of their syllable roles. 

The syllable shape constraint (18-25) shows first of all that
syllable roles are modelled as {\em types} under \code{self}. 

\startpiece
\begin{verbatim}
18 shape :=   
19  ( self:(nucleus & seg:vowel) &
20    left:self:onset 
21  ; self:(~nucleus) &
22      ( self:onset &  left:self:(~onset)
23      ; self:coda & left:self:(~coda)  
24      )          
25  ).
26 
27 constraints := syllabify & shape.
\end{verbatim}
\stoppiece
Lines (19-20) capture
the fact that syllable nuclei in MH are always vowels and that every
syllable nucleus is preceded by an onset. In (21-22) a non-nuclear
position that is an onset may only license preceding non-onsets,
thus disallowing complex onsets; similarly for codas in (23). In
(27) generic \code{syllabify} is intersected with \code{shape}, since segmental positions
must be prosodified {\em and} conform to language-specific
shape restrictions.

The constraints under (28-30), included for completeness, merely
ensure proper termination of segmental strings at the word periphery.

\startpiece
\begin{verbatim}
28 word := self:('+ini' & prom:up & onset).
29 end  := left:self:('+fin' & ~onset) &
30            self:'-fin'.
\end{verbatim}
\stoppiece
Prosodic prespecification (31-36) faithfully models what was stated 
in prose in section \ref{sec:DPM}. 

\startpiece
\begin{verbatim}
31 prosodic_prespecification :=
32    ( cat:((b2&(~(past;pres)));b3;b4) &
33      self:onset
34    ;
35      cat:(~((b2& (~(past;pres)));b3;b4))
36    ).
\end{verbatim}
\stoppiece
\noindent We proceed in (37-41) with a rudimentary definition of first (\code{v1})
and second (\code{v2}) stem vowel which is sufficient for our toy
fragment.

\startpiece
\begin{verbatim}
37 v1 := is(low) & cat:(past & b1;b7).
38 v1 := is(round & '-hi') & 
39         cat:(b1 & ~ past).
40 v2 := is(low).
41 v2 := is(front & '-hi').
\end{verbatim}
\stoppiece
The larger grammar mentioned above contains a full binary decision tree for
each vowel. Still, even here one can see the use of 
type formulae
like \code{round \& '-hi'} to classify segments phonologically.

Next come a number of exemplary inflectional affixes (42-79), again
simplified. The zero affixes (42-45, 47-54) are phonologically  just like the
zero alternant in (10) in taking up no segmental space.

\startpiece
\begin{verbatim}
42 %  initial "0" prefix
43 '#'(More) := More & self:'+ini' &
44  cat:(~ fut & ~ infinitive &
45        ( b1 ; (~ pres & (b3 ; b4)) )).
46 
47 %  final "0" suffix
48 '#'(More) := More &
49  self:'-ini' & left:self:'+fin' &
50  ( cat:(sg & masc & third & past) & 
51      left:left:is(~front)
52  ; cat:(sg & masc & third & pres) & 
53      left:left:is(front)
54  ).
55 
56 %  overt prefix
57 ji(More) := self:'+ini' &
58  obl(is(i),obl(is(i), More)) &
59  cat:(fut & third & ((sg&masc) ; pl) & 
60       (b1 ; b2)).
\end{verbatim}
\stoppiece
The segmental content of all other affixes is specified via possibly
repeated instances of \code{obl}, since affixes are
nonalternating. Apart from the respective \code{cat}egorial
information, positional type information 
\code{'+ini','+fin'} ensures that prefixes and suffixes
are properly restricted to wordinitial and wordfinal position. Note
that the glide-initial {\em ji-} prefix specifies an initial /i/ (58)
which will be prosodified as \code{onset} by means of
\code{syllabify}. This representational assumption is in
line with other recent work in phonological theory which 
standardly analyzes glides as nonsyllabic high vowels. Hence, even in
MH we have a case where segmental classes and prosodic roles
don't align perfectly. 

To control second stem vowel apophony, some suffixes demand (53,73)
or forbid (51) front vowels 
two positions to their left.

\startpiece
\begin{verbatim}
61 u(More) := obl(is(u)&self:'+fin',More)& 
62    left:left:is(~ (vowel & ~front)) & 
63    cat:(pl & ( (past & third) 
64          ; (fut & ~ first) )).
65 
66 a(More):=obl(is(a)&self:'+fin',More)& 
67    left:left:is(~ (vowel & ~front)) & 
68    cat:((past & third & sg & fem) 
69         ; (pres & sg & fem & b5)).
70 
71 et(More) := 
72  obl(is(e),obl(is(t)&self:'+fin',More))&
73  left:left:is(front) & 
74  cat:(pres & sg & fem & ~b5).
75 
76 im(More) := 
77  obl(is(i),obl(is(m)&self:'+fin',More))&
78  left:left:is(~ (vowel & ~front)) &
79  cat:(pres & pl & masc).
\end{verbatim}
\finalforcedpage
\stoppiece
\finalpagebreak

\noindent Others posit the weaker demand
\code{vowel} $\rightarrow$ \code{front} (62,67,78), thus not forbidding
consonantal fillings of the position adressed by \code{left:left}.

The \code{stem} definition (80-82) for a regular triliteral 
is parametrized
for the three root segments and the inflectional \code{Suffixes} to
follow.

\startpiece
\begin{verbatim}
80 stem(C1, C2, C3, Suffixes) := 
81  obl(is(C1),x_0(v1,obl(is(C2),
82  x_0(v2,obl(is(C3), Suffixes))))).
83 
84 affixes(Stem, '#'(end)) := '#'(Stem).
85 affixes(Stem, a(end)) := '#'(Stem).
86 affixes(Stem, et(end)) := '#'(Stem).
87 affixes(Stem, im(end)) := '#'(Stem).
88 affixes(Stem, u(end)) := yi(Stem).
89 
90 verbform([C1 & consonant,C2 & consonant,
91           C3 & consonant], Category) :=
92  root_letter_tree([C1,C2,C3]) & word &
93  affixes( prosodic_prespecification & 
94           stem(C1,C2,C3, Suffixes), 
95           Suffixes) &  cat:Category.
\end{verbatim}
\stoppiece
 Given the informal description in section \ref{sec:DPM}, the
succession of obligatory root and alternating stem vowel positions
now looks familiar. It should be obvious how to
devise analogous stem definitions for quadriliterals (e.g. {\em
  mi\nichi\esh ev}) and cluster verbs (e.g. {\em flirtet}).

A rather simple  tabulation of \code{affixes} lists (a subset of) the allowable prefix-suffix
cooccurrences in the MH verbal paradigm (84-88) before everything is
put together in the definition for \code{verbform}, parametrized for
a list of root segments and \code{Category} (90-95). Note how
\code{prosodic\_prespecification} is intersected with \code{stem} in
(93-94), exploiting the power of the description level to restrict stem
realizations without diacritical marking of stem vs affix domains on
the object level. The subgoal \code{root\_letter\_tree} (92) will be discussed below.

When proving a goal like \code{verbform([g,m,r],
  b1\&\/third\&\/pl\&\/fut)},
the MicroCUF interpreter will enumerate the set of all candidate result 
feature structures, including one that describes the grammatical
surface string {\em  jigmeru}. An implementation of the IOP, to be described next, must
therefore complement the setup established sofar to exclude the suboptimal
candidates. While the subtle intertwining of zero alternant preference
and constraint solving described above has its theoretical merits, a much
simpler practical solution was devised. In a first step, the small
finite set of all candidate solutions for a goal is collected, together with numerical
`disharmony' values representing each candidate's degree of optimality.
Disharmony is defined as the binary number that results from application
of the mapping \{\code{unmarked} $\mapsto$ 01$_2$, \code{marked} $\mapsto$
10$_2$\} to the left-to-right markedness vector of a segmental string:
e.g., j$_{01}$i$_{01}$g$_{01}$a$_{10}$m$_{01}$r$_{01}$u$_{01}$ yields the disharmony value
01010110010101$_2$ = 5525$_{10}$ $>$ 5477$_{10}$ = 01010101100101$_2$ from
j$_{01}$i$_{01}$g$_{01}$m$_{01}$e$_{10}$r$_{01}$u$_{01}$. 
Step two is a straightforward search for the candidate(s) with
minimal disharmony.
\section{Parsing and generation}
\label{sec:pargen}
The preceding paragraph described how to compute surface
forms given roots and categories. However, this
generation procedure amounts to an 
inefficient {\em generate-and-minimize}
mechanism which must compute otherwise useless suboptimal candidates as a
byproduct of optimization. More importantly, due to the
nonmonotonicity of optimization it is not obvious how to invert the
procedure for {\em efficient parsing} in order to
derive root and category given a surface form. 

A first solution which comes to mind is to implement parsing as {\em analysis-by-synthesis}.
A goal like \code{ParseString\&\/verbform(Root,Category)} is submitted
to a first run of the MicroCUF constraint solver, resulting in instantiations for \code{Root}
and \code{Category} iff a proof consistent with the grammar was found. With these
instantiations, a second run of MicroCUF uses the full {\em 
  generate-and-minimize} mechanism to compute {\em optimal} strings 
\code{OptString1}, \dots,\code{OptStringN}. The parse is accepted iff \code{ParseString\&%
\/(OptString1;\dots;OptStringN)} is consistent. Note that for this
solution to be feasible it is essential that constraints are inviolable,
hence their evaluation in the first run can disregard optimization.
The main drawbacks of {\em analysis-by-synthesis} are that two runs
are required 
and that the inefficiencies of {\em generate-and-minimize} are not avoided. 

The new solution recognizes the fact that
bidirectional processing of DPM would be easy without optimization.
We therefore seek to perform all optimization at compile time.
The idea is this: exploiting the finiteness of natural language paradigms we
compute  -- using {\em  generate-and-minimize} --  each paradigm cell of
e.g. the verbal paradigm of MH for a suitable root. However, while
doing so we record the proof sequence of relational clause invocations
employed in the derivation of each optimal form, using the fact that
each clause has a unique index 
in internal representation. Such proof sequences have two
noteworthy properties. By definition they first of all record
just clause applications, therefore naturally abstracting over all
non-relational parameter fillings of top-level goals. In particular, proving
a goal like \code{verbform([g,m,r], b1;b2)} normally looses the information associated with the
root and category parameters in the proof sequence representation
(although these parameters {\em could} indirectly influence the proof if
relationally encoded choices in the grammar were dependent on
it). Secondly, we can profitably view each 
proof sequence as a linear finite state automaton (FSA$_{cell}$). Since a
paradigm is the union of all its cells, a complete abstract paradigm
can therefore be represented by a unique minimal deterministic
FSA$_{para}$ which is computed as
the union of all FSA$_{cell}$ followed by determinization and
minimization. 
At runtime we just need to run FSA$_{para}$ as a {\em finite-state
  oracle} in parallel with the MicroCUF constraint solver. This means
that each proof step that uses a clause $k$ must be sanctioned by
a corresponding $k$-labelled FSA transition. 
With this technique  we are now able to
efficiently restrict the search space to just the optimal proofs;  the
need for run-time optimization in 
DPM processing has been removed. 
However, a slight caveat is necessary: to apply the technique it must be possible to partition
the data set 
into a finite number of equivalence classes. This condition is
e.g. automatically fulfilled for all phenomena which exhibit a paradigm structure. 

What are the possible advantages of this hybrid FSA-guided constraint
processing technique? First of all, it enables a particularly simple
treatment of {\em unknown words} for root-and-pattern morphologies,
surely a necessity in the face of ever-incomplete lexicons. If the grammar is set up properly
to abstract from segmental detail of the \code{Root}
segments as much as possible, then these details are also absent in the proof
sequences. Hence a single FSA$_{para}$ merging these sequences in
effect represents an abstract paradigm which can be used for a large number of
concrete instantiations. We thus have a principled way of
parsing words that contain roots not listed in the lexicon. 
However, we want the system not to overgenerate, mistakenly analyzing known
roots as unknown. Rather, the system should return the semantics of
known roots and also respect their verbal class affiliations as well as other
idiosyncratic properties. This is the purpose of
the \code{root\_letter\_tree} clauses in
(96-123). 
\finalpagebreak

\startpiece
\begin{verbatim}
 96 root_letter_tree([g|Rest]) := 
 97  root_letter_tree_g(Rest).
 98 root_letter_tree([~g|_]) := 
 99  cat:sem:'UNKNOWN'.
100  
101 root_letter_tree_g([m|Rest]) := 
102  root_letter_tree_gm(Rest).
103 root_letter_tree_g([d|Rest]) := 
104  root_letter_tree_gd(Rest).
105 root_letter_tree_g([~m&~d|_]) := 
106  cat:sem:'UNKNOWN'.
107  
108 root_letter_tree_gm([r]) := 
109  cat:(b1 & sem:'FINISH' 
110      ; b2 & sem:'BE FINISHED').
111 root_letter_tree_gm([~r|_]) := 
112  cat:sem:'UNKNOWN'.
113 root_letter_tree_gd([r]) := 
114  cat:( b1 & sem:'ENCLOSE' 
115      ; b2 & sem:'BE ENCLOSED' 
116      ; b3 & sem:'FENCE IN' 
117      ; b4 & sem:'BE FENCED IN' 
118      ; b5 & sem:'DEFINE' 
119      ; b6 & sem:'BE DEFINED' 
120      ; b7 & sem:'EXCEL' 
121      ).
122 root_letter_tree_gd([~r|_]) := 
123  cat:sem:'UNKNOWN'.
\end{verbatim}
\stoppiece
For each level in the letter tree a new terminal branch is added that covers the
complement of all attested root segments at that level
(99,106,112,123). This terminal branch is assigned an \code{'UNKNOWN'}
semantics, whereas known terminal branches record a proper semantics
and categorial restrictions.
During off-line creation of the
proof sequences we now simply let the system backtrack over all choices in
the \code{root\_letter\_tree} by feeding it a totally underspecified
\code{Root} parameter. The resulting FSA$_{para}$ represents both the
derivations of  all known roots and of all possible unknown root types covered by
the grammar. While this treatment results in a homogenous
grammar integrating lexical and grammatical aspects, it considerably enlarges
FSA$_{para}$. It might therefore be worthwhile to separate lexical
access from the grammar, running a separate proof of
\code{root\_letter\_tree(Root)} to enforce root-specific restrictions
{\em after} parsing with the abstract paradigm alone. It remains to be seen
which approach is more promising w.r.t. overall space and time efficiency.

A second advantage of separating FSA guidance from constraint
processing, as compared to pure finite-state transducer approaches,
is that we are free to build sufficient expressivity into the
constraint language. For example it seems that one needs token identity,
i.e. structure sharing, in phonology to cover instances of antigemination,
assimilation, dissimilation and reduplication in an insightful way. It
is wellknown that token identity is not finite-state representable and cumbersome to emulate in
practice (cf. \citeNP[157]{antworth:90} on a FST attempt at
reduplication vs the DPM treatment of infixal reduplication in
Tigrinya verbs described in \citeNP[238-247]{walther:97}). Also, it
would be fascinating to extend the constraint-based 
approach to phonetics. However, a pilot study
reported in \citeN{walther.kroeger:94a} has found it necessary to use
arithmetic constraints to do so, again transcending finite-state power.
Finally, to the extent that sign-based approaches to grammar like HPSG
are on the right track, the smooth integration of phonology and
morphology arguably is better achieved within a uniform formal basis
such as MicroCUF which is expressive enough to cover the recursive
aspects of syntax and semantics as well.

In conclusion, some notes on the pilot implementation. The
MicroCUF system was modified to produce two new incarnations of the
MicroCUF interpreter, one to record proof sequences, the other to perform FSA-guided
proofs. FSA$_{para}$ was created with the help of finite-state tools
from AT\&T's freely available {\em fsm} package
({\small\tt http://www. research. att. com /sw /tools /fsm/}). I have
measured speedups of more than 10$^2$ for the generation of MH
forms ($<$ 1 second with the new technique), although parse times in
the range of 1 \dots 4 seconds on a Pentium 200 MHz PC with 64 MByte
indicate that the current prototype is still too slow by a factor of more than 10$^2$. 
However, there is ample room for future improvements. Besides drawing from
the wealth of optimizations found in the logic programming literature
to generally accelerate MicroCUF (e.g., term encoding of feature
structures, memoization) we can also analyze the internal structure of
FSA$_{para}$ to 
gain some specific advantages. This is due to the fact that each maximal linear sub-FSA
of length $k > 1$ corresponds to a deterministic proof subsequence whose
clauses should be partially executable at compile time, subsequently
saving $k-1$ proof steps at runtime.
\section{Conclusion}
\label{sec:disc}
This paper has described a computational, declarative
approach to prosodic morphology which uses inviolable constraints
formulated in a sufficiently expressive formalism (here: MicroCUF) together
with a restrictive incremental optimization component.
The approach has been illustrated by implementing an a-templatic
 analysis of a fragment of Modern Hebrew verbs. The full grammar
 behind the illustrative fragment covers additional detail such as
 antigemination effects ({\em noded-im, *no\underline{dd}-im} `they
 (masc.) wander'), spirantization, B7 sibilant metathesis, etc. Also,
 the formalization of X/\lingnull\/ presented here is actually a
 special case of the more powerful notion of {\em resequencing}, whose 
 application to Tigrinya vowel coalescence and metathesis was
 demonstrated in \citeN{walther:97}.

Despite the initial emphasis on incremental optimization, a compilation
technique was later proposed 
to remove the need for run-time
optimization and guarantee fully bidirectional processing of prosodic morphology.
Although the general idea of using a finite-state oracle to guide a parser has
been previously proposed for context-free grammars 
\cite{johnson:96},
both the details of our implementation of the idea and its specific application to prosodic
morphology are believed to be novel. It was emphasized how the
proposed technique aided in a simple treatment of unknown words.
Note that unknown words are not normally integrated into finite-state transducer
models of prosodic morphology, although the necessary extensions
appear to be possible (K. Beesley, p.c.). Finally, the fact that a
hybrid setup rather than a pure finite-state approach was chosen has
been motivated {\em inter alia} by reference to additional phenomena such as
antigemination and reduplication that require the richer notion of
token identity. Future research will especially focus on detailed
analyses of reduplication phenomena to secure the relevance of the
present approach to prosodic morphology at large.

\small\label{sec:references}

\end{document}